\documentclass[10pt,twocolumn,letterpaper]{article}

\usepackage{times}
\usepackage{epsfig}
\usepackage{graphicx}
\usepackage{amsmath}
\usepackage{amssymb}
\usepackage[numbers]{natbib}
\usepackage{float}
\usepackage{tabularx, booktabs}
\usepackage{url}
\usepackage{xcolor}
\usepackage{framed,multirow}

\usepackage[breaklinks=true,bookmarks=false]{hyperref}

\begin{document}

%%%%%%%%% TITLE
\title{Patch-based adaptive weighting with segmentation\\ and scale (PAWSS) for visual tracking}

\author{Xiaofei Du$^1$ \\ xiaofei.du.13@ucl.ac.uk \and
	Alessio Dore$^2$ \\ alessio.dore@deliveroo.co.uk \and
	Danail Stoyanov$^1$ \\danail.stoyanov@ucl.ac.uk
}
\date{
	$^1$ Surgical Vision Group, University College London, UK \\
	$^2$ Deliveroo, London, UK}

%{\tt\small ganchuang1990@gmail.com}

\maketitle
%\thispagestyle{empty}

%%%%%%%%% ABSTRACT
\begin{abstract}
Tracking-by-detection algorithms are widely used for visual tracking, where the problem is treated as a classification task where an object model is updated over time using online learning techniques. In challenging conditions where an object undergoes deformation or scale variations, the update step is prone to include background information in the model appearance or to lack the ability to estimate the scale change, which degrades the performance of the classifier. In this paper, we incorporate a Patch-based Adaptive Weighting with Segmentation and Scale (PAWSS) tracking framework that tackles both the scale and background problems. A simple but effective colour-based segmentation model is used to suppress background information and multi-scale samples are extracted to enrich the training pool, which allows the tracker to handle both incremental and abrupt scale variations between frames. Experimentally, we evaluate our approach on the online tracking benchmark (OTB) dataset and Visual Object Tracking (VOT) challenge datasets. The results show that our approach outperforms recent state-of-the-art trackers, and it especially improves the successful rate score on the OTB dataset, while on the VOT datasets, PAWSS ranks among the top trackers while operating at real-time frame rates.
\end{abstract}

\section{Introduction}
\label{sec:intro}
Tracking-by-detection is one of the highly successful paradigms for visual object tracking \citep{hare2011struck,kalal2012tracking,kim2015sowp,wang2015visual}. A typical tracking-by-detection algorithm treats the tracking problem as a classification task, it begins with a detector initialized with a bounding region in the first frame, and updates the detection model over time with collected positive and negative samples. The choice of the samples used to update the classifier is critical for robust tracking and maintaining the model's reliability but as the object moves background information within the bounding box is falsely included in the sample descriptors which causes corruption in the classifier. Additionally, real world objects usually undergo different transformations, such as deformation, scale change, occlusion or all at the same time, which render the robust estimation of scale difficult.

To address these problems, different methods have been proposed to decrease the effects of background information in the model template, such as using patch-based descriptors and assigning weights based on the pixel spatial location or appearance similarity \citep{comaniciu2003kernel,he2013visual,lee2014visual}. Directly integrating a segmentation step into the tracking update has also been effective \citep{duffner2013pixeltrack,godec2013hough}. In this paper, we follow a similar idea to incorporate a Patch-based Adaptive Weighting with Segmentation and Scale (PAWSS) into the tracking framework. It uses a simple but effective colour-based segmentation model to assign weights to the patch-based descriptor which decreases background information influences within the bounding box, and also a two-level sampling strategy is introduced to extract multi-scale samples, which enables the tracker to handle both incremental and abrupt scale variations between frames. Our method is evaluated and compared with the state-of-the-art methods on the online tracking benchmark (OTB)~\citep{wu2013online} and VOT challenge datasets with promising results demonstrating that PAWSS is among the best performing real-time trackers without any specific code optimisation.

\section{Proposed Algorithm}
\label{sec:method}
\subsection{Probabilistic Segmentation Model for Patch Weighting}
\label{sec:weight}
We used the patch-based descriptor to represent the appearance of the object. In frame~$t$, the bounding box~$\Omega$ is evenly decomposed into~$n_\varphi$ non-overlapping patches~$\{\varphi_i\}_{i=1:n_\varphi}$, then the descriptor~$\vec{\Phi}_{\Omega, t}$ is constructed by concatenating the low-level feature vectors of all the patches in their spatial order. Since background information is potentially included in the bounding box, we would like to incorporate an global probabilistic segmentation model~\citep{collins2005online,duffner2013pixeltrack} to assign weights to the patches based on their colour appearance. 
\begin{equation}
\vec{\Phi}_{\Omega, t} = [w_{1,t}\vec{\phi}_1^T,\dots,w_{n_\varphi,t}\vec{\phi}_{n_\varphi}^T]^T
\end{equation}
where~$w_i$ is the weight of the feature vector $\vec{\phi}_i$ of the $i$-th patch~$\varphi_i$.
The global segmentation model is based on colour histogram by using a recursive Bayesian formulation to discriminate foreground and background.

Let~$y_{1:t}$ be the colour observation of a pixel from frame~$1$ to~$t$, the foreground probability of that pixel at frame~$t$ is based on the tracked results from previous frames
\begin{equation}
\begin{split}
p(c_t=1|y_{1:t})=Z^{-1}\sum_{c_{t-1}}&p(y_t|c_t=1)p(c_t=1|c_{t-1})\\
&p(c_{t-1}|y_{1:t-1}) 
%&p(c_t=0|c_{t-1})=0.6 \qquad p(c_t=0|c_{t-1})=0.4
\label{equ:foreground_prob}
\end{split}
\end{equation}
where~$c_{t}$ is the class of the pixel at frame~$t$: 0 for background, and 1 for foreground, and~$Z$ is a normalization constant to keep the probabilities sum to 1. The transition probabilities for foreground and background~$p(c_t|c_{t-1})$ where $c \in \{0, 1\}$ are empirical choices as in~\citep{duffner2013pixeltrack}. The foreground histogram~$p(y_t|c_t=1)$ and the background histogram~$p(y_t|c_t=0)$ are initialized from the pixels inside the bounding box and from those which are surrounding the bounding box (with some margin between) in the first frame, respectively. For the following frames, the colour histogram distributions are updated using the tracked result.
\begin{equation}
\begin{split}
p(y_t|c_t=1)=&\delta p(y_t|y_t\in \Omega_t)) \\
&+(1-\delta)p(y_{t-1}|c_{t-1}=1)  
\label{equ:colour_hist}
\end{split}
\end{equation}
where~$0\le\delta\le1$ is the model update factor. $\Omega_t$ represents the tracked bounding box in frame~$t$. Instead of treating every pixel equal, the weighting of a pixel also depends on the patch where it is located. Patches with higher weight are more likely to contain object pixels and vice versa. So the colour histogram update for colour observation $y_t$ of current frame~$t$ is defined as  
\begin{equation}
p(y_t|y_t\in \Omega_t) = \frac{\sum_{i=1}^{n_\varphi}w_{i,t-1} N_{y_t\in \varphi_{i,t}}}{\sum_{i=1}^{n_\varphi} w_{i,t-1}\sum_{x_t} N_{x_t\in \varphi_{i,t}}}
\end{equation}
where~$N_{y_t\in \varphi_{i,t}}$ represents the number of pixels with colour observation $y_t$ in the~$i$-th patch~$\varphi_{i,t}$ in frame~$t$, and~$x_t$ represents any colour observation in frame~$t$, so the denominator means the weighted number of all the pixel colour observations in the bounding box~$\Omega_t$.

The weights~$w_{i,1}$ for all the patches are initialized as 1 at the first frame, and then are updated based on the segmentation model
\begin{align}
\label{equ:multi}
& w_{i,t}=\delta \bar{w}_{i,t} +(1-\delta)w_{i,t-1}  \\
& \bar{w}_{i,t} = \frac{\varpi_{i,t}}{\max_{1 \leq i \leq n_\varphi} \varpi_{i,t}} \\
& \varpi_{i,t}=\frac{\sum_{x_t}p(x_t|c_t=1)N_{x_t\in \varphi_{i,t}}}{\sum_{x_t}N_{x_t\in \varphi_{i,t}}}
\end{align}
where $\varpi_{i,t}$ denotes the average foreground probability of all pixels in the patch~$\varphi_{i,t}$ in the current frame~$t$, it is normalized so the highest weight update~$\bar{w}_{i,t}$ equals 1. The patch weight~$w_{i,t}$ is then updated gradually over time. We omit the background probability~$p(c_t=0|y_{1:t})$ since it is similar to Eq.~\ref{equ:foreground_prob}.

\begin{figure}[tb]
	\centering
	\includegraphics[width=0.5\textwidth]{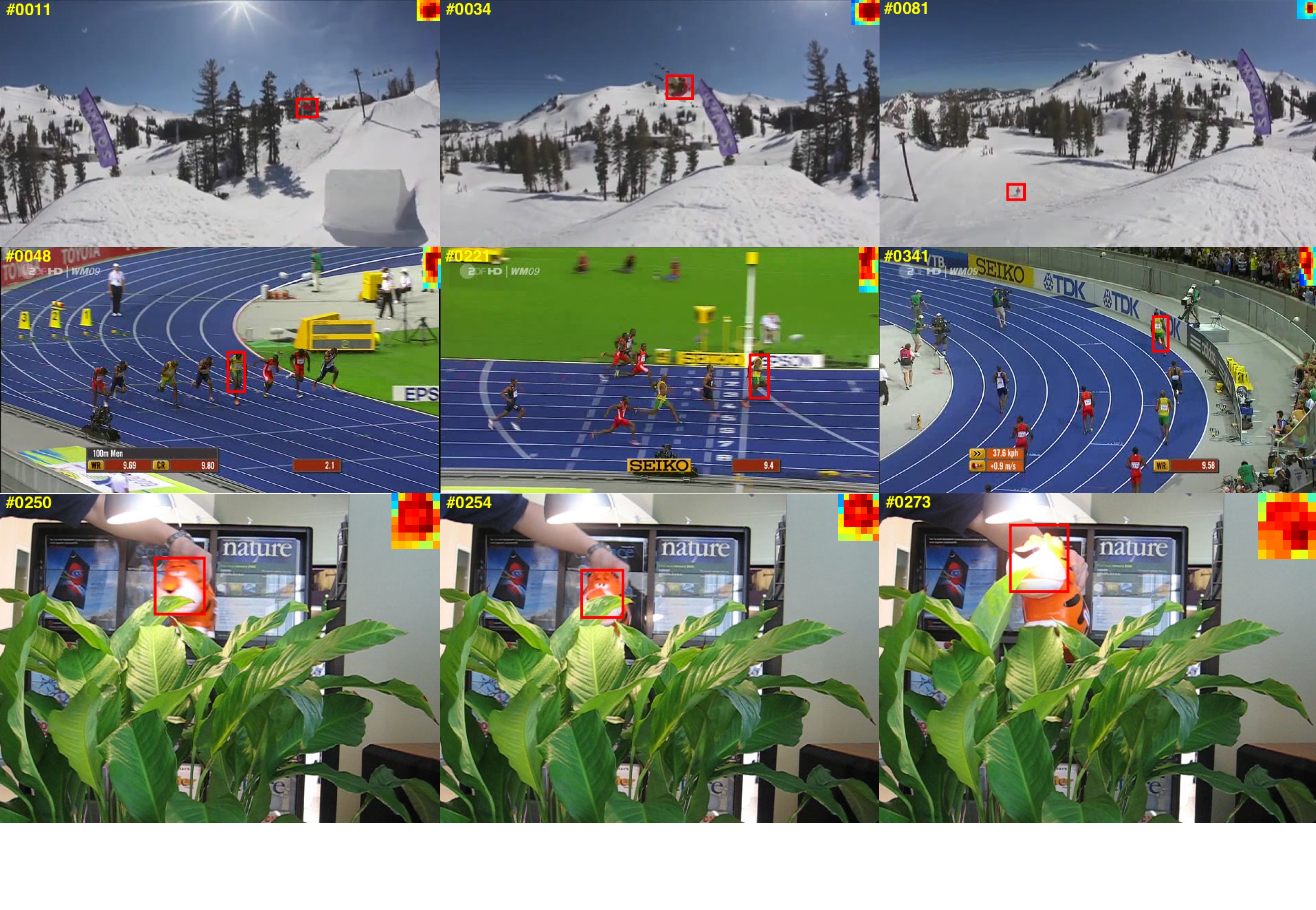}
	\caption{Example patch weights are shown for the highlighted bounding box displayed in the top corner of the image. Warmer colour indicates higher foreground possibility.}
	\label{fig:patch_weight}
\end{figure}

Unlike the weighting strategy in \citep{chen2013constructing,kim2015sowp} by analysing the similarities between neighbouring patches, our patch weighting method is simple and straightforward to implement, the weight update for each patch is independent from each other, and only relies on the colour histogram based segmentation model. We show examples of the patch weight evolvement in Figure~\ref{fig:patch_weight}. The patch weight thumbnails are displayed on the top corner of each frame, which indicate the objectness in the bounding box and also reflect the deformation of the object over time.
Since we update the segmentation model based on the previous patch weight, and in turn the segmentation model facilitates updating the weight patches. This co-training strategy enhances the weight contrast between foreground and occluded patches, which suppresses the background information efficiently.

\subsection{Scale Estimation}
\label{sec:scale}
\begin{figure}[tb]
	\centering
	\includegraphics[width=0.5\textwidth]{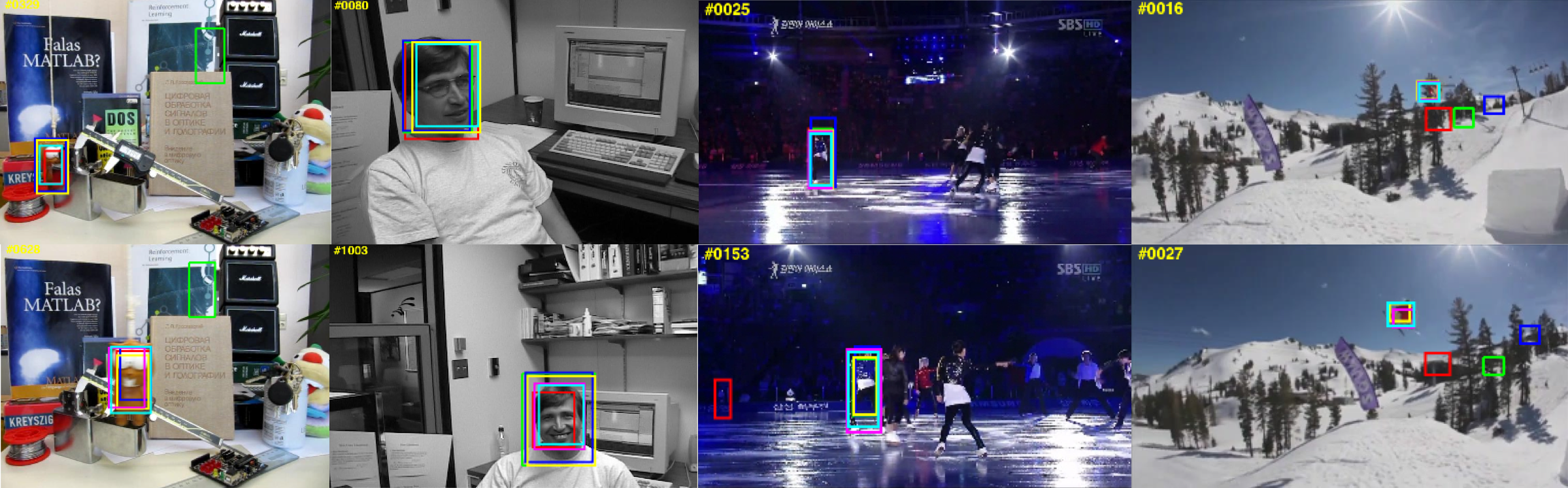}
	\caption{Examples from of objects undergo challenging transformations for tracking, inclusion of background information or partial object within the bounding box usually degrade the classifier.}
	\label{fig:example}
\end{figure}

The tracked object often undergoes complicated transformations during tracking, for example, deformation, scale variations, occlusion et. al as shown in Figure~\ref{fig:example}. Fixed-scale bounding box estimation is ill-equipped to capture the accurate extents of the object, which would degrade the classifier performance by providing samples which are either partial cropped or include background information.

%In the original Struck~\citep{hare2011struck} tracking framework,
When locating the object in a new frame, all the bounding box candidates are collected within a searching window, and the bounding box with the maximum classification score is selected to update the object location. Rather than making a suboptimal decision by choosing from fixed-scale samples, we augment the training sample pool with multi-scale candidates. Obviously, the scales of the augmented samples are critical. We consider two complementary strategies that handle both incremental and abrupt scale variations.

Firstly, to deal with relatively small scale changes between frames, we build a scale set~$S_r$
\begin{equation}
S_r = \{s|s=\lambda^m s_{t-1} \} \quad m\in[-\frac{n_r-1}{2},\dots, \frac{n_r-1}{2} ]
\label{equ:scale}
\end{equation} 
where~$\lambda$ is a fixed value which is slightly larger than $1.0$. It is set to accurately search the scale change. $n_r$ is the scale number in the scale set $S_r$.~$s_{t-1}$ is the scale of the object in frame~$t-1$ compared with the initial bounding box in the first frame. Considering object scale usually does not vary too much between frames, scale set~$S_r$ includes scales which are close to the previous frame.

Secondly, when object undergoes abrupt scale changes between frames, scale set~$S_r$ is unable to keep pace with the speed of the scale variations. To address this problem, we build an additional scale set~$S_p$ by incorporating Lucas-Kanade tracker (KLT) \citep{bouguet2001pyramidal,shi1994good}, which helps us estimate the scale change explicitly. We randomly pick $n_{pt}$ points from each patch in the bounding box $\Omega_{t-1}$ of frame~$t-1$, and tracked all these points in the next frame~$t$. With sufficient well-tracked points, we can estimate the scale variation between frames by comparing the distance changes of the tracked point pairs. 

\begin{figure}[tb]
	\centering
	\includegraphics[width=0.5\textwidth]{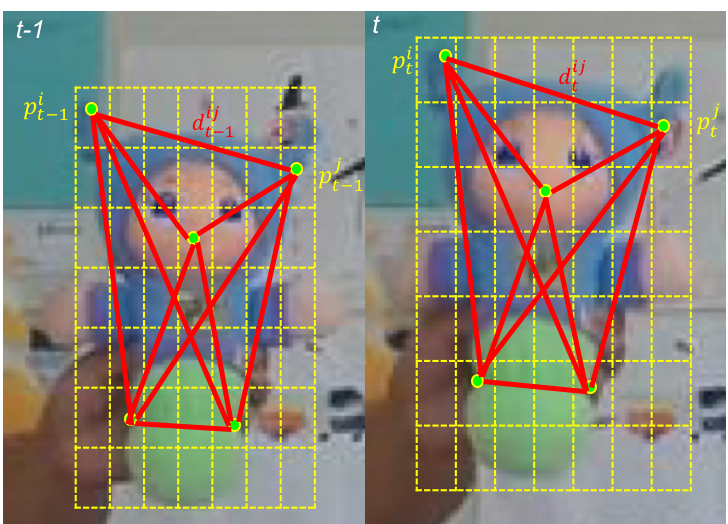}
	\caption{Illustration of the scale estimation by using the KLT tracker. Random points located on the patches are picked in frame~$t-1$, and are tracked in the next frame~$t$ by the KLT tracker, the distance ratio of point pairs~$(p^{i}, p^{j})$ between two frames are used for scale estimation.}
	\label{fig:scale}
\end{figure}

We illustrated the scale estimation by KLT tracker in Figure~\ref{fig:scale}.
Let $p_{t-1}^{i}$ denotes one picked point in the previous frame $t-1$ and its matched point~$p_{t}^{i}$ in the current frame~$t$. We compute the distance~$d_{t-1}^{ij}$ between point-pair~$(p_{t-1}^{i}, p_{t-1}^{j})$, and the distance~$d_{t}^{ij}$ between the matched point-pair~$(p_{t}^{i}, p_{t}^j)$.
For all the matched point pairs, we compute the distance ratio between the two frame
\begin{equation}
V = \{s | s = d_{t}^{ij}/d_{t-1}^{ij}\} \quad i\neq j 
\end{equation}
where $V$ is the set with all the distance ratios. We sort~$V$ by value and pick the median element $s_p=V_{sorted}(\frac{n}{2})$ as the potential scale change of the object. To make the scale estimation more robust, we uniformly sample the scales ranging between~$[1, s_p]$ or $[s_p, 1]$ to construct the scale set $S_p$.
\begin{equation}
S_p = \{ s|s=1+i\frac{s_p-1}{n_p-1} \} \quad 0 \le i < n_p
\end{equation}
where $n_p$ is the scale number in the scale set $S_p$. When the object is out-of-view, occluded or abruptly deforms, the ratio of well-tracked points will be low. In that case, the estimation from the KLT tracker will be unreliable. In our implementation, when the ratio is lower than 0.5, we then set $s_p=1$, therefore the scale set $S_p$ will only add samples with the previous scale into the candidate pool. Only when there are enough points well tracked, the estimation from the KLT tracker will be trusted. We fuse these two complementary scale sets $S_r$ and $S_p$ into $S_f = S_r \cup S_p$ to enrich our sample candidate pool. To show the effectiveness, we evaluate our proposed tracker in section~\ref{sec:res} with or without scale set $S_p$ estimated by the KLT tracker. 
\subsection{Tracking Framework}
\label{sec:framework}
We incorporate PAWSS into the Struck~\citep{hare2011struck}. The algorithm relies on an online structured output SVM learning framework which integrates the learning and tracking. It directly predicts the location displacement between frame, avoiding the heuristic intermediate step for assigning binary labels to training samples, which acheives top performance in the OTB dataset~\citep{wu2013online}.

Given the bounding box $\Omega_{t-1}$ in the previous frame~$t-1$, sample candidates are extracted in a searching window~$r_w$, which centers at the $\Omega_{t-1}$ in the current frame~$t$, unlike other tracking-by-detection approaches, we adapt a two-level sampling statergy. On the first level, all the bounding box samples are extracted with fixed-scale $s_{t-1}$, on the second level, multi-scale samples are extracted to enrich the sample pool.

First, the searching window is chosen at the same as above centered at the $\Omega_{t-1}$ with a radius of $r_w$, since we have the second level to make the final decision, rather than extracting sample per pixel, we extract samples at a down-sample factor of 2, which could decrease the candidate number by 4, then the weighted patch-based descriptor of each candidate is constructed, and we select the bounding box with the maximum classification score not as the final decision, but as the search center for our second level. After this step, the rough location of the object is narrowed into a smaller area. Like discussed in Section~\ref{sec:scale}, given the scale $s_{t-1}$ in the previous frame~$t-1$, to handle small scale variation between frames, we construct the scale set~$S_r$, which includes scales which are close to $s_{t-1}$. Additionally, to deal with potential abrupt scale changes, we randomly pick $n_{pt}$ points from each patch of the bounding box $\Omega_{t-1}$, and pass all these points to the KLT tracker to generate the scale set $S_p$. This scale set is estimated explicitly by the KLT tracker and facilitates to augment the scale estimation. The two scale sets $S_r$ and $S_p$ are complementary to handle different scenarios. Then we use the fused scale set $S_f$ to extract bounding box candidates. We set a smaller search window with search radius of $r_s$, centering at the bounding box selected in the first level, and we construct multiple candidates for each pixel within the search window. The scales of candidates at one pixel are set as scales in the fused scale set $S_f$. We then evaluate all the multi-scale samples, and select the bounding box sample with the maxiumn score as the final location of the object. For multiple bounding box samples with the same scores, the sample whose scale is closer to 1.0 is selected to prevent potential gradual shrinking or enlargement of the bounding box. 

Then, the classifier, the colour-based segmentation model and the weights of all patches are updated as discussed in Section~\ref{sec:weight}. Finally, the whole process starts at the next frame. Additionally, to prevent introducing potential corrupt samples to the classifier, the classifier only updates when the similarity between the tracked object and the positive support vectors are above certain threshold~$\eta$.
\section{Results}
\label{sec:res}
\textbf{Implementation Details} \quad
Our algorithm is publicly available online\footnote{\url{https://github.com/surgical-vision/PAWSS}} and is implemented in C++ and performs at about $7$ frames per second with an i7-2.5GHz CPU without any optimisation. For structured output SVM, we are using a linear kernel and the parameters are empirically set as $\delta = 0.1$ in Eq.~\ref{equ:colour_hist} and Eq.~\ref{equ:multi}, $\lambda=1.003$ in Eq.~\ref{equ:scale}, the scale numbers of the scale set are $n_r=n_p=11$. The number of extracted points from each patch $n_{pt}=5$. The updating threshold for the classifier is set as $\eta=0.3$. For each sequence, we scale the frame to make sure the minimum side length of the bounding box is larger than 32 pixels, and the search window radius $r_w$ is fixed to~$(W+H)/2$, where~$W$ and $H$ represents the width and height of the scaled bounding box, respectively, and the search window radius $r_s$ is fixed to 5 pixels.
Selecting the right features to describe the object appearance plays a critical role to differentiate object and background. We tested different low-level features and found that the combination of HSV colour and gradient features achieves the best results. The patch number affects the tracking performance, too many patches increase the computation and too less patches do not robustly reflect the local appearance of the object. We tested different patch numbers, and selected $n_\varphi=49$ to strike a performance balance. 

\subsection{Online Tracking Benchmark (OTB)}
The OTB dataset~\citep{wu2013online} includes $50$ sequences tagged with $11$ attributes, which represent the challenging aspects for tracking such as illumination variation, occlusion, deformation et al. The tracking performance is quantitatively evaluated using both precision rate (PR) and success rate (SR), as defined in~\citep{wu2013online}. PR/SR scores are depicted using precision plot and success plot, respectively. The precision plot shows the percentage of frames whose tracked centre is within certain Euclidean distance (20 pixels) from the centre of the ground truth. Success plot computes the percentage of frames whose intersection over union overlap with the ground truth annotation is within a threshold varying between 0 and 1, and the area under curve (AUC) is used for SR score. 
To evaluate the effectiveness of incorporating the scale set proposed by the KLT tracker, we provide two versions of our tracker as PAWSSa and PAWSSb: PAWSSa only includes scale set~$S_r$, while PAWSSb includes both $S_r$ and $S_p$ for scale estimation.

\begin{figure}[tb]
	\centering
	\includegraphics[width=0.5\textwidth]{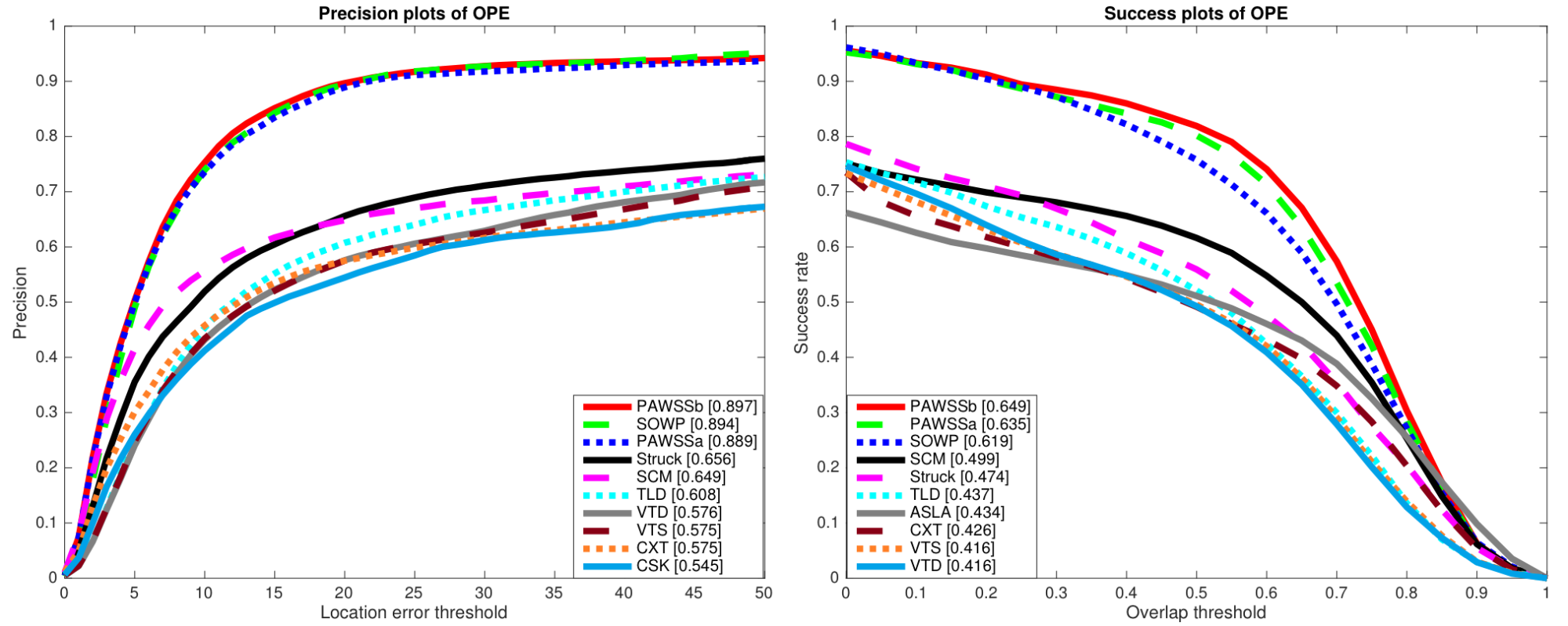}
	\caption{Comparison of the precision and success plots on the OTB with the top 10 trackers; the PR scores are illustrated with the threshold at 20 pixels and the SR scores with the average overlap (AUC) in the legend.}
	\label{fig:plot}
\end{figure}

We use the evaluation toolkit provided by Wu~\citep{wu2013online} to generate the precision and success plots for the one pass evaluation (OPE) of the top 10 algorithms in Figure~\ref{fig:plot}. The toolkit includes 29 benchmark trackers, besides that we also include SOWP tracker. It is shown that PAWSSb achieves the best PR/SR scores among all the trackers. For a more detailed evaluation, we also compared our tracker with the state-of-the-art trackers in Table~\ref{table:attr2}. Notice that in all the attribute field, our tracker achieves either the best or the second best PR/SR scores. Our tracker achieves 36.7\% gain in PR and 36.9\% gain in SR over Struck~\citep{hare2011struck}. By using a simple patch weighting strategy and training with adaptive scale samples, the performance shows that our tracker provides comparable PR scores, and higher SR score compared with SOWP \citep{kim2015sowp}. PAWSSa tracker improves the SR score by 2.6\% considering gradually small changes between frames, PAWSSb improves the SR score by 4.8\% by incorporating scales estimated by the external KLT tracker. Specifically, when the object undergoes scare variation PAWSS achieves a performance gain of 10.3\% in SR over SOWP.

\begin{table*}
	\caption{Comparison of the PR/SR score with state-of-the-art trackers in the OPE based on the 11 sequence attributes: illumination variation (IV), scale variation (SV), occlusion (OCC), deformation (DEF), motion blur (MB), fast motion (FM), in-plane rotation (IPR),  out-of-plane rotation (OPR), out-of-view (OV), background cluttered (BC) and low resolution (LR). The best and the second best results are shown in {\color{red} \textbf{red}} and {\color{blue}\textbf{blue}} colours respectively.}
	\label{table:attr2}
	\begin{center}
		\resizebox{\textwidth}{!}
		{
			\begin{tabular}{c||c|c|c|c|c|c|c|c|c}
				\hline
				& Struck~\citep{hare2011struck}
				& DSST~\citep{danelljan2014accurate} & SAMF~\citep{li2014scale} &FCNT~\citep{wang2015visual} & MTA~\citep{lee2015multihypothesis} 
				&MEEM~\citep{zhang2014meem} 
				&SOWP~\citep{kim2015sowp} & PAWSSa & PAWSSb\\
				%	\noalign{\smallskip}
				\hline    													  
				IV(25) & 0.558 / 0.428 & 0.727 / 0.534& 0.735 / 0.563& 0.830 / 0.598& 0.738 / 0.547& 0.778 / 0.548& 0.842 / 0.596& {\color{blue} \textbf{0.860 / 0.616}} & {\color{red} \textbf{0.880 / 0.648}}\\
				SV(28) & 0.639 / 0.425 & 0.723 / 0.516& 0.730 / 0.541& {0.830} / 0.558 & 0.721 / 0.478&  0.809 / 0.506& {\color{red}\textbf{0.849}} / 0.523& {\color{red}\textbf{0.849}} / {\color{blue}\textbf{0.564}} & {\color{red} \textbf{0.849 / 0.577}}\\
				OCC(29) & 0.564 / 0.413 & 0.845 / {\color{blue}\textbf{0.619}}& 0.716 / 0.534& 0.797 / 0.571& 0.772 / 0.563& 0.815 / 0.560& {\color{blue}\textbf{0.867}} / 0.603& 0.859 / 0.618 & {\color{red} \textbf{0.872 / 0.634}}\\
				DEF(19) & 0.521 / 0.393 & 0.813 / 0.622& 0.660 / 0.510&  0.917 / 0.644&  0.851 / 0.622& 0.859 / 0.582& {\color{blue}\textbf{0.918 / 0.666}}& 0.908 / 0.656 & {\color{red} \textbf{0.934 / 0.688}}\\
				MB(12) & 0.551 / 0.433 & 0.651 / 0.519& 0.547 / 0.464& {\color{red}\textbf{0.789}} / 0.580& 0.695 / 0.540& 0.740 / 0.565& 0.716 / 0.567&   {\color{blue}\textbf{0.786}} / {\color{blue}\textbf{0.593}} & 0.783 / {\color{red} \textbf{0.603}}\\
				FM(17) & 0.604 / 0.462  & 0.663 / 0.515& 0.517 / 0.435& 0.767 / 0.565&  0.677 / 0.524& 0.757 / 0.568& 0.744 / {\color{blue}\textbf{0.575}}& {\color{blue}\textbf{0.784}} / 0.572 & {\color{red} \textbf{0.792 / 0.587}}\\
				IPR(31) & 0.617 / 0.444 & 0.691 / 0.507& 0.765 / 0.560& 0.811 / 0.555&  0.773 / 0.547 & 0.810 / 0.531& 0.847 / 0.584& {\color{red}\textbf{0.860}} / {\color{blue} \textbf{0.594}} & {\color{blue} \textbf{0.852}} / {\color{red} \textbf{ 0.600}}\\
				OPR(39) & 0.597 / 0.432 & 0.763 / 0.554& 0.733 / 0.535& 0.831 / 0.581& 0.777 / 0.557& 0.854 / 0.566&  0.896 / 0.615 & {\color{blue}\textbf{0.898 / 0.623}} & {\color{red} \textbf{0.901 / 0.635}}\\
				OV(6) & 0.539 / 0.459  & 0.708 / 0.609& 0.515 / 0.459& 0.741 / 0.592& 0.612 / 0.534& 0.730 / 0.597& {\color{blue}\textbf{0.802 / 0.635}}&  0.771 / 0.611 & {\color{red} \textbf{0.828 / 0.645}}\\
				BC(21) & 0.585 / 0.458  & 0.708 / 0.524& 0.694 / 0.517& 0.799 / 0.564& 0.795 / 0.592& 0.808 / 0.578&  0.839 / 0.618 & {\color{blue}\textbf{0.847 / 0.632}} & {\color{red} \textbf{0.859 / 0.647}}\\
				LR(4) & 0.545 / 0.372  & 0.459 / 0.361& 0.497 / 0.409& {\color{red}\textbf{0.765 / 0.514}}& 0.579 / 0.397& 0.494 / 0.367& 0.606 / 0.410&  {\color{blue}\textbf{0.679 / 0.504}} & 0.669 / 0.500\\
				\hline
				Avg.(50)& 0.656 / 0.474 &  0.777 / 0.570& 0.737 / 0.554& 0.856 / 0.599& 0.812 / 0.583& 0.840 / 0.570& {\color{blue}\textbf{0.894}} / 0.619&  0.889 / {\color{blue}\textbf{ 0.635}} & {\color{red} \textbf{0.897 / 0.649}}\\
				\hline
			\end{tabular}
		}
	\end{center}	
\end{table*}

\begin{figure}[tb]
	\centering
	\includegraphics[width=0.5\textwidth]{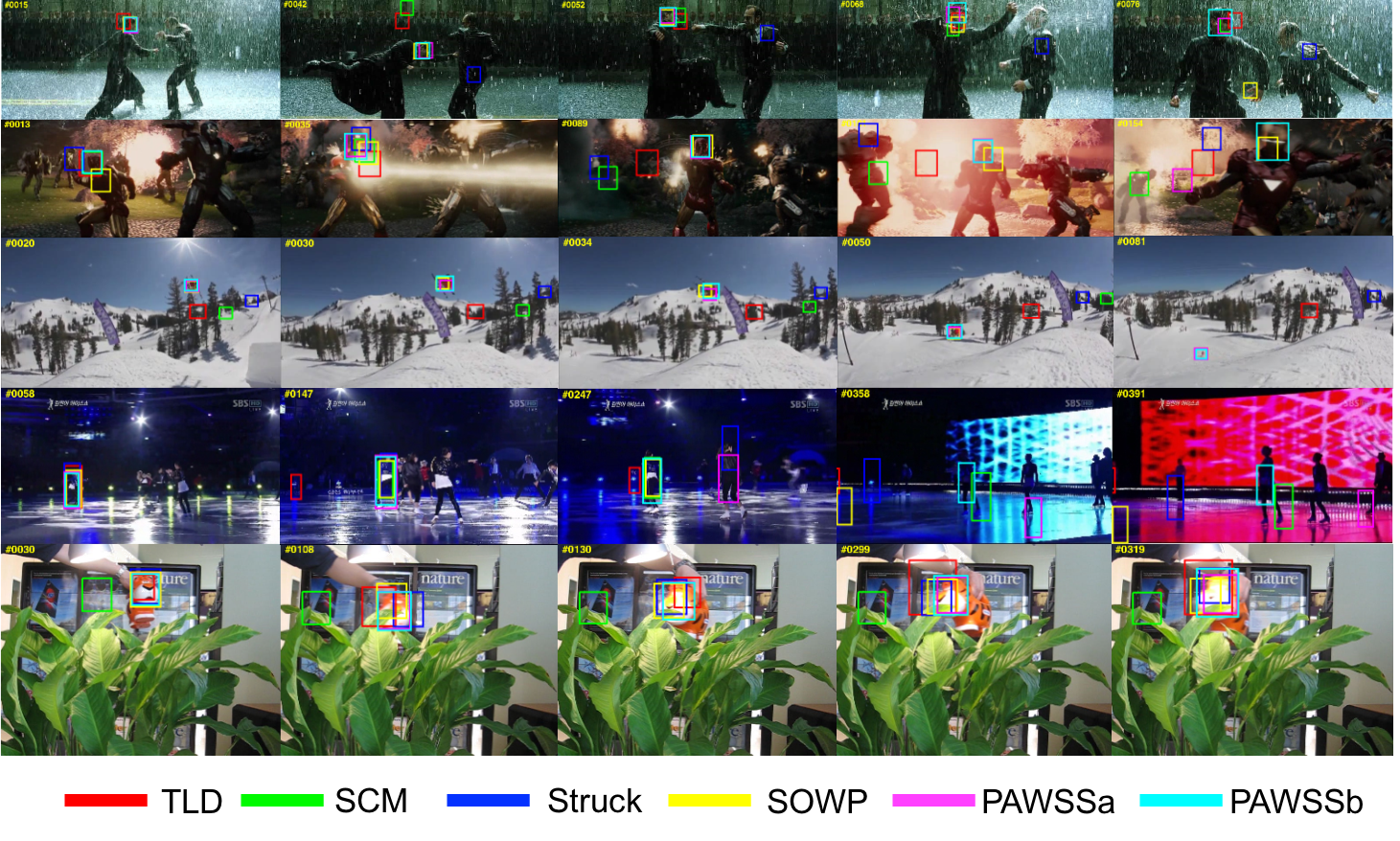}
	\caption{Comparison of the tracking results of our proposed tracker PAWSS with SOWP~\citep{kim2015sowp} and three conventional trackers: TLD~\citep{kalal2012tracking}, SCM~\citep{zhong2012robust} and Struck~\citep{hare2011struck} on some especially challenging sequences in the benchmark.}
	\label{fig:seq_c}
\end{figure}

\begin{figure}[tb]
	\centering
	\includegraphics[width=0.5\textwidth]{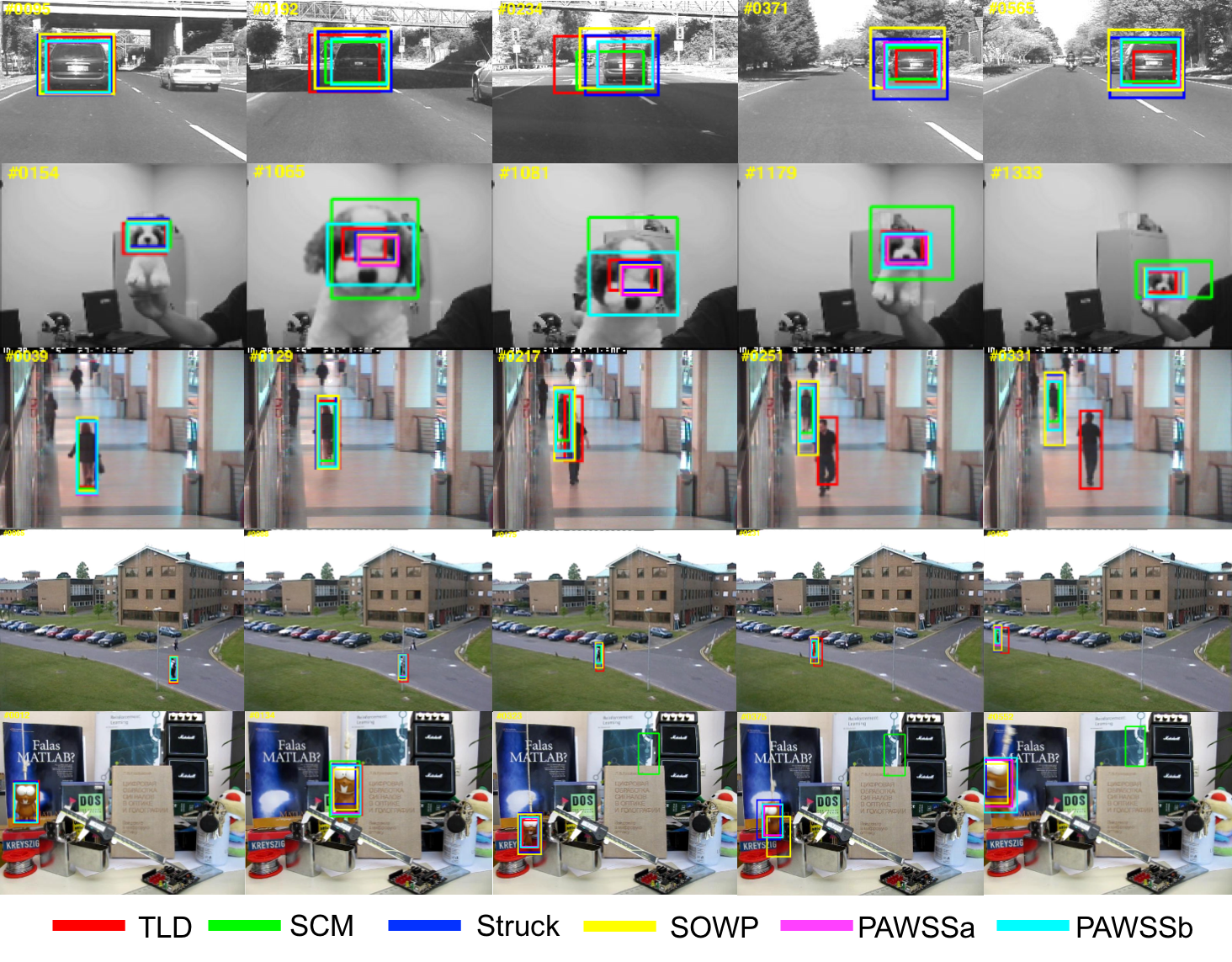}
	\caption{Comparison of the tracking results of our proposed tracker PAWSS with SOWP~\citep{kim2015sowp} and three conventional trackers: TLD~\citep{kalal2012tracking}, SCM~\citep{zhong2012robust} and Struck~\citep{hare2011struck} on some sequences with scale variations in the benchmark.}
	\label{fig:seq_s}
\end{figure}

We show tracking results in Figure~\ref{fig:seq_c} and Figure~\ref{fig:seq_s} with the top trackers including TLD~\citep{kalal2012tracking}, SCM~\citep{zhong2012robust}, Struck~\citep{hare2011struck}, SOWP~\citep{kim2015sowp} and the proposed PAWSSa and PAWSSb. In Figure~$\ref{fig:seq_c}$, five challenging sequences are selected from the benchmark dataset, which include illumination variation, scale variations, deformation, occlusion or background clusters. PAWSS can adapt when the object deforms in a complicated scene and track the target accurately.
In Figure \ref{fig:seq_s}, we select five representative sequences with different scale variations. PAWSS can well track the object with scale variation, while other trackers drift away. The results show that our proposed tracking framework PAWSS can track the object robustly through sequence by using the weighting strategy to suppress the background information within the bounding box, and also by incorporating scale estimation allowing the classifier to train with adaptive scale samples. Please see the supplementary video for more sequence tracking results.

\subsection{Visual Object Tracking (VOT) Challenges}
For completeness, we also validated our algorithm on VOT2014 (25 sequences) and VOT2015 (60 sequences) datasets. VOT datasets use ranking-based evaluation methodology: accuracy and robustness. Similar to SR rate for OTB dataset, the accuracy measures the overlap of the predicted result and the ground truth bounding box, while the robustness measures how many times the tracker fails during tracking. A failure is indicated whenever the tracker loses the target object which means the overlap becomes zero, and it will be re-initialized afterwards. All the trackers are evaluated, compared and ranked based on with respect to each measure separately using the official evaluation toolkit from the challenge \footnote{\url{http://www.votchallenge.net/}}.

\begin{figure}[tb]
	\centering
	\includegraphics[width=0.5\textwidth]{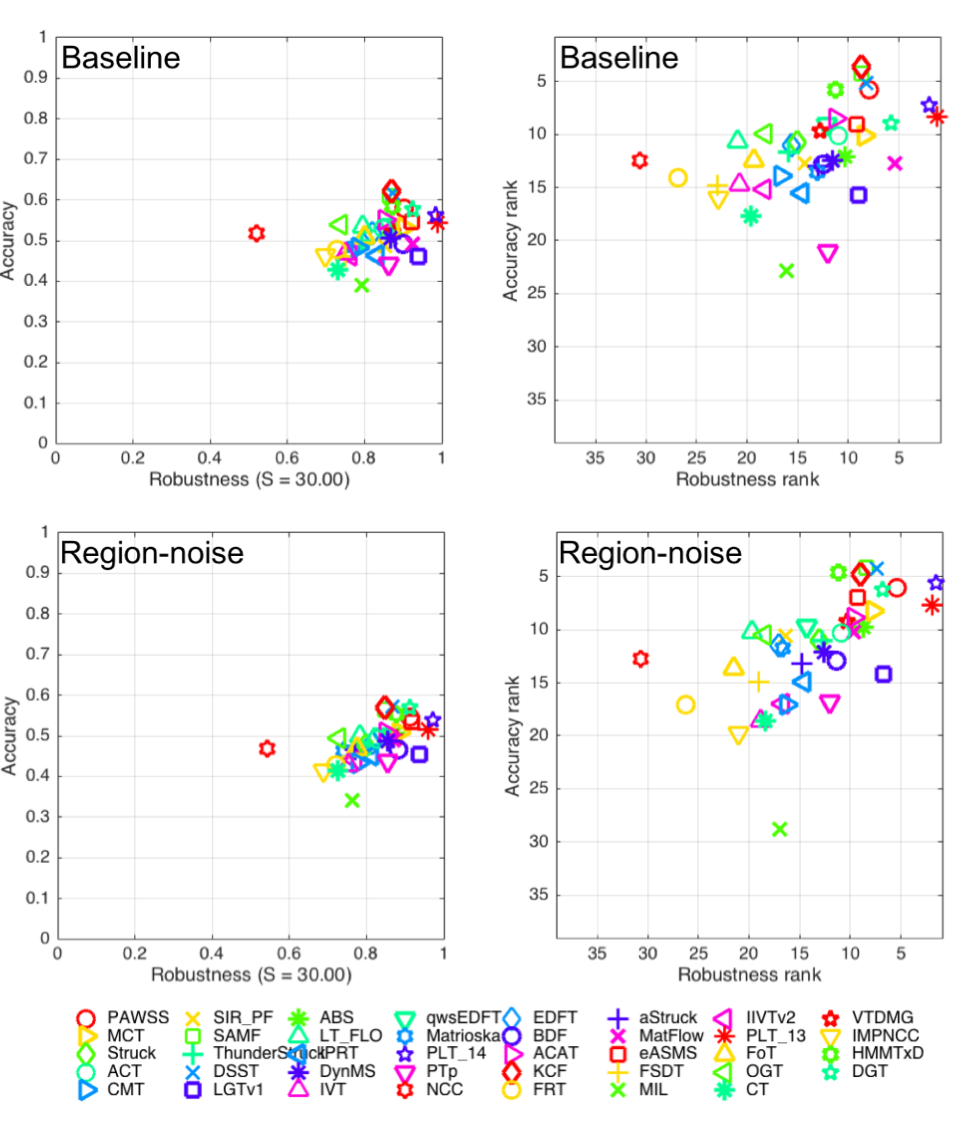}
	\caption{The accuracy-robustness score and ranking plots with respect to the baseline and region-noise experiments of VOT2014 dataset. Tracker is better if its result is closer to the top-right corner of the plot.}
	\label{fig:vot2014}
\end{figure}

\bigskip
\noindent\textbf{VOT2014} \quad
The VOT2014 challenge includes two experiments: baseline experiment and region-noise experiment. In baseline experiment, a tracker runs on all the sequences by initializing with the ground truth bounding box on the first frame; while in the region-noise experiment, the tracker is initialized with a random noisy bounding box with the perturbation in the 10\% of the ground truth bounding box size.~\citep{Kristan2015}.
The ranking plots with 38 trackers are shown in Figure \ref{fig:vot2014} for comparing PAWSS with the top three trackers: DSST~\citep{danelljan2014accurate}, SAMF~\citep{li2014scale}, KCF~\citep{henriques2015high} in Table~\ref{table:vot2014}. For both the experiments our PAWSS has lower accuracy score~$0.58/0.55$, but less failures ~$0.88/0.78$ and have a second average rank. But considering the tracking process of the experiments: once a failure is detected, the tracker will be re-initialized, to eliminate the effect of achieving higher accuracy score by more re-initialization steps, we performed experiments without the re-initialization, also shown in Table~\ref{table:vot2014}. The results show that PAWSS has the highest accuracy score~$0.51/0.48$ among all the trackers without re-initialization, which means it is more robust than the other trackers.

\begin{table*}
	\caption{The results of VOT2014 baseline and region-noise experiments with and without-re-initialization. The best and the second best results are shown in {\color{red} \textbf{red}} and {\color{blue}\textbf{blue}} colours respectively.}
	\label{table:vot2014}
	\begin{center}
		\resizebox{\textwidth}{!}
		{
		\begin{tabular}{c||cc|cc|c|cc|cc|c|c}
		\hline
		\multirow{3}{*}{} & \multicolumn{5}{|c|}{\textbf{Baseline}} & \multicolumn{5}{|c|}{\textbf{Region-noise}} & \multirow{3}{*}{Avg Rank}\\\cline{2-11}
		&\multicolumn{2}{|c|}{Accuracy} 
		&\multicolumn{2}{|c|}{Robustness} & Accuracy (w/o)
		&\multicolumn{2}{|c|}{Accuracy} 
		&\multicolumn{2}{|c|}{Robustness} & Accuracy (w/o)  \\\cline{2-11}
		& Score & Rank & Failure & Rank& Score
		& Score & Rank & Failure & Rank& Score  \\\hline
		DSST~\citep{danelljan2014accurate} & 0.62 & 5.16 & \color{blue}1.16 & \color{blue} 8.2 & 0.47 & \color{blue}0.57 & \color{blue} 4.32 & \color{blue}1.28 & \color{blue}7.4 & 0.43 & \color{red}6.27 \\\hline	
		SAMF~\citep{li2014scale} & \color{blue}0.61 & \color{blue} 4.32 & 1.28 & 8.68 & \color{blue}0.50 & \color{red}0.57 & \color{red}{4.2} & 1.43 & 8.44 & \color{red}0.48 & 6.41 \\\hline	
		KCF~\citep{henriques2015high} & \color{red}0.62 &\color{red}{3.68} & 1.32 & 8.68 & 0.40 & 0.57 & 4.84 & 1.51 & 9.00 & 0.36 & 6.92 \\\hline	
		PAWSSb & 0.58 & 5.80 & \color{red}0.88 & \color{red}{8.00} & \color{red}0.51 & 0.55 & 6.08 & \color{red}0.78 & \color{red}{5.4} & \color{red}0.48 & \color{blue} 6.32 \\\hline	 
		\end{tabular}
		}
	\end{center}
\end{table*}

\begin{figure}[tb]
	\centering
	\includegraphics[width=0.5\textwidth]{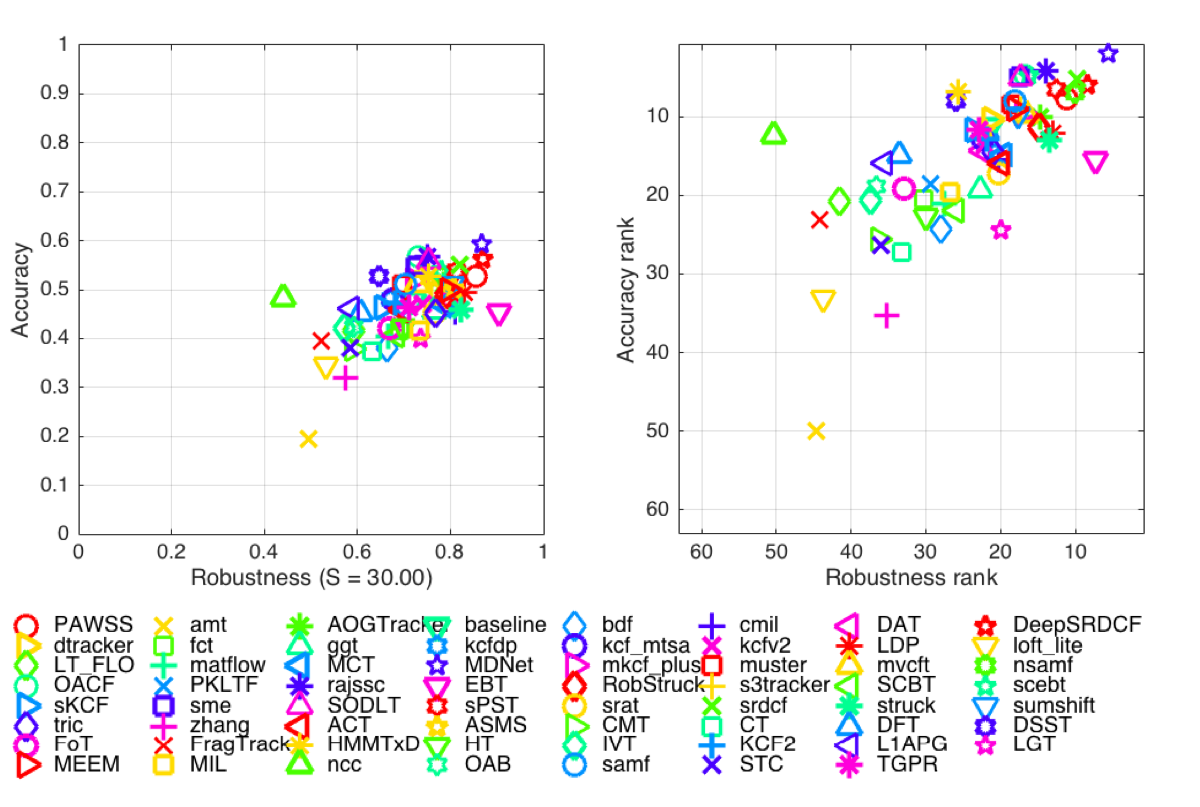}
	\caption{The accuracy-robustness ranking plots of VOT2015 dataset. Tracker is better if its result is closer to the top-right corner of the plot.}
	\label{fig:vot2015_ranking}
\end{figure}

\bigskip
\noindent\textbf{VOT2015} \quad
Finally, we evaluated and compared PAWSS with 62 trackers on the VOT2015 dataset. The VOT2015 challenge only includes baseline experiment, and the ranking plots are shown in Figure~\ref{fig:vot2015_ranking}. 
In VOT2015~\citep{kristan2015visual}, expected average overlap measure is introduced which combines both per-frame accuracies and failures in a principled manner. Compared with the average rank used in VOT2014, expected overlap has a more clear practical interpretation.
We list the score / rank and expected overlap of the top trackers from VOT2015 \citep{kristan2015visual} which are either quite robust or accurate, the above VOT2014 top three trackers DSST~\citep{danelljan2014accurate}, SAMF~\citep{li2014scale}, KCF~\citep{henriques2015high}\footnote{This is an improved version of the original tracker.}, and the baseline NCC tracker in Table~\ref{table:vot2015}. It can be shown that the average rank is not always consistent with the expected overlap. 
Our tracker PAWSS is among those top trackers (ranks the~$7$-th), also PAWSS achieves better than any of the VOT2014 top trackers on VOT2015 dataset.

\begin{table}[H]
	\caption{VOT2015 score/ranking and expected overlap results from the top trackers of VOT2014, VOT2015 and the baseline tracker. The NCC tracker is the VOT2015 baseline tracker. Trackers marked with $\dagger$ are submitted to VOT2015 without publication.}
	\label{table:vot2015}
	\begin{center}
		\resizebox{0.5\textwidth}{!}
		{
			\begin{tabular}{c||cc|cc|c|c}
				\hline
				\multirow{3}{*}{} & \multicolumn{4}{|c|}{\textbf{Baseline}} & \multirow{3}{*}{Avg Rank} & \multirow{3}{*}{Exp Overlap}\\\cline{2-5}
				&\multicolumn{2}{|c|}{Accuracy} & \multicolumn{2}{|c|}{Robustness} & & \\\cline{2-5}
				& Score & Rank & Failure & Rank& &   \\\hline
				MDNet~\citep{nam2015learning} & 0.59 & 2.03 & 0.77 & 5.68 & 3.86 & 0.378 \\\hline	
				DeepSRDCF~\citep{Danelljan_2015_ICCV_Workshops} & 0.56 & 5.92 & 1.00 &8.38 & 7.15 & 0.318 \\\hline
				EBT~\citep{wang2014ensemble} & 0.45 & 15.48 & 0.81 & 7.23  & 11.36& 0.313\\\hline
				SRDCT~\citep{danelljan2015learning} & 0.55 & 5.25 & 1.18 &9.83 & 7.54&0.288  \\\hline
				LDP~\citep{lukevzivc2016deformable} & 0.49 & 12.08 & 1.30 &13.07  & 12.58&0.279 \\\hline	
				sPST~\citep{hua2015online} & 0.54 & 6.57 & 1.42 & 12.57  & 9.57& 0.277\\\hline
				\textbf{PAWSSb} & \textbf{0.53} & \textbf{7.75}  &\textbf{ 1.28} & \textbf{11.22} &\textbf{9.49} & \textbf{0.266} \\\hline	
				NSAMF$\dagger$  & 0.53 & 7.02 & 1.45 &10.1 & 8.56 & 0.254 \\\hline
				RAJSSC~\citep{Zhang_2015_ICCV_Workshops} & 0.57 & 4.23 & 1.75 & 13.87 &9.05& 0.242\\\hline
				RobStruck$\dagger$  & 0.49 & 11.45 & 1.58 &14.82&13.14 &0.220 \\\hline
				DSST~\citep{danelljan2014accurate} &0.53  & 8.05  & 2.72 &26.02 & 17.04 &0.172 \\\hline
				SAMF~\citep{li2014scale} &0.51  & 7.98  & 2.08 &18.08 & 13.03 &0.202 \\\hline
				KCF~\citep{henriques2015high} &0.47  & 12.83  & 2.43 &21.85 & 17.34 &0.171 \\\hline
				NCC* &0.48  & 12.47  & 8.18 &50.33 & 31.4 &0.080 \\\hline 
			\end{tabular}
		}
	\end{center}
\end{table}

\section{Conclusions}
\label{sec:conclusions}
In this paper, we propose a tracking-by-detection framework, called PAWSS, for online object tracking. It uses a colour-based segmentation model to suppress background information by assigning weights to the patch-wise descriptor. We incorporate scale estimation into the framework, allowing the tracker to handle both incremental and abrupt scale variations between frames. 
The learning component in our framework is based on Struck, but we would like to point out that theoretically our proposed method can also support other online learning techniques with effective background suppression and scale adaption.
The performance of our tracker is thoroughly evaluated on the OTB, VOT2014 and VOT2015 datasets and compared with recent state-of-the-art trackers. Results demonstrate that PAWSS achieves the best performance in both PR and SR in the OPE for OTB dataset. It outperforms Struck by $36.7\%$ and $36.9\%$ in PR/SR scores. Also, it provides a comparable PR score, and improves SR score by $4.8\%$ over SOWP. On the VOT2014 and VOT2015 datasets, PAWSS has relatively lower accuracies but the lowest failure rate among the top trackers, we evaluated without re-initialization, and achieves the highest performance. 

\section*{Acknowledgements}
Xiaofei Du is supported by the China Scholarship Council (CSC) scholarship. The work has been carried out as part of an internship at Wirewax Ltd, London, UK. The work was supported by the EPSRC (EP/N013220/1, EP/N022750/1, EP/N027078/1, NS/A000027/1, EP/P012841/1), The Wellcome Trust (WT101957, 201080/Z/16/Z) and the EU-Horizon2020 project EndoVESPA (H2020-ICT-2015-688592).

\bibliographystyle{IEEEtran}
\bibliography{lib}

\end{document}